\documentclass{article}
\usepackage[final]{neurips_2020}
\usepackage[utf8]{inputenc} 
\usepackage[T1]{fontenc}    
\usepackage{hyperref}       
\usepackage{url}            
\usepackage{booktabs}       
\usepackage{amsfonts}       
\usepackage{nicefrac}       
\usepackage{microtype}      
\usepackage{graphicx}
\usepackage{algorithm,algpseudocode}
\usepackage{amsmath}
\usepackage{amsthm}
\usepackage{mathtools}
\usepackage{amssymb}
\usepackage{pdfpages}
\graphicspath{ {./images/} }
\title{Tree DNN: A Deep Container Network}

\author{%
  Brijraj Singh \\
  Data Science\\
  Sony Research India\\
  \texttt{brijraj.08@gmail.com} \And   
  Swati Gupta\\
  Department of Computer Science\\
  JIIT, Noida, India\\
  \texttt{swatigupta.iitr@gmail.com} \And   
    Mayukh Das \\
  Microsoft Research, India \\
  \texttt{mayukhdas@microsoft.com} \\
    \And   
    Praveen Doreswamy Naidu \\
  On Device AI\\
  Samsung Research Institute, Bangalore, India\\
  \texttt{praveen.dn@samsung.com} \And   
  Sharan Kumar Allur \\
  On Device AI\\
  Samsung Research Institute, Bangalore, India\\
  \texttt{sharan.allur@samsung.com} \\
}
\begin{document}
\maketitle
\begin{abstract}
Multi-Task Learning (MTL) has shown its importance at user products for fast training, data efficiency, reduced overfitting etc. MTL achieves it by sharing the network parameters and training a network for multiple tasks simultaneously. However, MTL does not provide the solution, if each task needs training from a different dataset. In order to solve the stated problem, we have proposed an architecture named TreeDNN along with it's training methodology. TreeDNN helps in training the model with multiple datasets simultaneously, where each branch of the tree may need a different training dataset. We have shown in the results that TreeDNN provides competitive performance with the advantage of reduced ROM requirement for parameter storage and increased responsiveness of the system by loading only specific branch at inference time.     
\end{abstract}

\section{Introduction}
Multitask learning (MTL) has proved its importance by facilitating the execution of multiple tasks  with just one DNN (Deep Neural Network) model (\cite{sandler2018mobilenetv2}). MTL is achieved by exploiting the generalized features extracted from the back-bone module of the model where each task is differentiated at the last layers of the DNN network. In general, MTL works well if all the tasks are related to each other where generic features are exploited by the backbone layers and task-specific features are extracted with the help of classification layers as explained by \cite{lu2022mtl}.
\par
In many commercial systems, it is often required to load multiple DNN models on working memory to perform sequential tasks. Application of camera models in mobile phones is one of the use-cases, where a user first goes with the basic camera and then switches among multiple options like night-mode, selfie, depth, sports, food, make-up etc. All categories of cameras use their respective DNN models for their corresponding tasks. A point to notice here is that, in general, all tasks are mutually exclusive and only one is performed at a time. As either a camera will work for the portrait features or it will work for night mode. As functional requirements are different hence, the training to their corresponding DNN model needs to be different as well. Therefore, an input sample should be plugged-in each DNN model exclusively unlike a typical MTL where same input is considered for multiple tasks.
\par
In order to solve the stated problem, we are proposing here a deep container model named as TreeDNN. TreeDNN can be considered as inverted tree, which has a trunk and multiple branches (one branch for each task). A TreeDNN is deployed to support multiple tasks requirement where each task needs training from different dataset. To train a TreeDNN model, we need to perform a joint training of all the branches followed by specialized training of each branch individually. A joint training is performed with the help of federated minibatch, which is prepared by considering all the datasets. The federated minibatch is inferenced through the trunk and all the branches of tree. Loss values are calculated at each branch and are aggregated for cumulative loss. In this work we have considered MobileNet-V2 (\cite{sandler2018mobilenetv2}, \cite{singh2019shunt}) as a backbone architecture.The cumulative loss is back-propagated through the branches and trunk of the tree to provide the generalized training to the model. Afterwards, the specialized training is given to each branch of the tree individually.   
\par
This work is proposed for the camera system of mobile phones, where multiple cameras are switched frequently. TreeDNN helps in reducing the model latency thereby improving the system responsiveness. It requires only 50\% of the total number of parameters for the storage requirement along with competitive predicting performance. 
\subsection{Related work}
MTL provides the facility to solve multiple tasks using one DNN model. Few MTL studies share parameters among their tasks and are known as shared trunk methodology as proposed by \cite{lu2022mtl} \cite{xie2022end} \cite{ma2018modeling}. All the MTL methods work in a similar manner, where input sample is inferenced through multiple operations and operations are shared among tasks as proposed by \cite{Ott_2022_WACV}. This way, it reduces efforts in training a task from scratch by utilizing the trunk as shown by \cite{santiago2021self}. There are multiple applications like \cite{keceli2022deep} proposed a MTL-based framework for identifying each plant's type and disease by considering raw image and transformed feature-maps from another pretrained DNN. 
\par
However, MTL methods do not provide solution, if each task needs training from a different dataset.
\begin{figure}[t]
  \centering
  \includegraphics[scale=0.3]{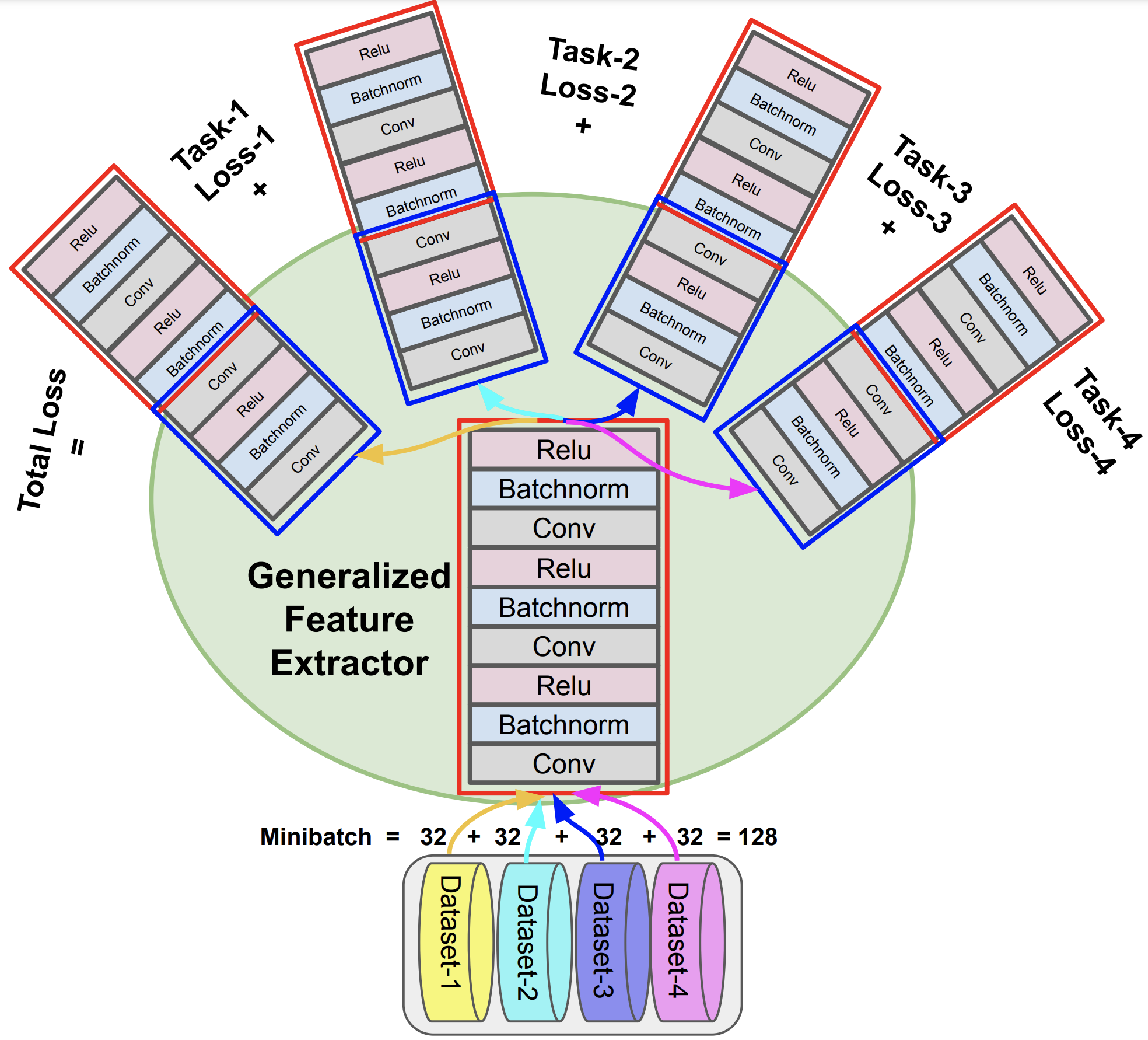}
  \caption{Training of TreeDNN}
  \label{Train_TreeDNN}
\end{figure}

\subsection{Proposed Solution}
TreeDNN is  a container network that is a collection of multiple DNN models prepared to serve the tasks independently. As the name suggests a portion of the DNN model is shared by all the tasks and is referred as trunk of the tree as shown in Figure \ref{Train_TreeDNN}. Each branch of the tree is dedicated to a particular task. In this work, TreeDNN utilizes the ops of vision model. Trunk of the tree is designed to receive generalized training through all the datasets. TreeDNN is designed to speed up the switching among the tasks, which are different branches of the tree therefore, branches of the tree need to be lighter (on scale of parameters). The speedup is attained because of keeping trunk of the tree on the working memory and consuming the time only in loading the particular branch of the tree. Whole process can be segregated among a) Model designing b) Dataset preparation and c) Training.
\par
\textbf{(a) Model Designing:}
A typical vision model is a combination of operations like Convolution, BatchNorm, Relu, Pooling, Linear, Fully connected layers etc. A TreeDNN architecture contains a trunk, branches and both are designed with the above mentioned vision operations. In this work, we have considered MobileNet-V2 as a backbone architecture therefore trunk of the Tree is mainly a collection of inverted residual blocks. Branch of the TreeDNN is prepared from the second half portion of Mobilenet-V2 and the number of layers are reduced/ increased based on the complexity of the task performed at that particular branch. The process of model creation is defined through line 25-30, in Algorithm 1. It is interesting to realize that $M_i:Trunk \rightarrow{Branch_i}$ is a fully-functioning DNN model capable to serve $Task_i$. 
\begin{algorithm}[h!]
\begin{small}
\caption{\textsc{TREE\_DNN($D_1, D_2, \ldots, D_k$)}}
\label{alg:treednnalgo}
\begin{algorithmic}[1]
\State $\vec{\mathbb{M}} \gets$ \Call{Model\_creation}{$Trunk, Branch_1, Branch_2, \ldots, Branch_k$}
\Procedure{Federated\_Batch\_Preparation}{$D_1, D_2, \ldots, D_k$}
\For{ $i \in (0, len(D_1)/B)$} \Comment{$Batch\_size= B$}
\State $Fed\_Batch_i= D_1[i*(B/k): (i+1)*B/k]|| \ldots ||D_k[i*(B/k): (i+1)*B/k]$
\EndFor
\EndProcedure
\Procedure{Generalized Training}{$\vec{\mathbb{D}}, Max\_epoch$}
\For{$epoch \in Max\_epoch$}
\For{$i \in k$} \Comment{Loss from each branch}
\For{$Fed\_Batch \in \vec{\mathbb{D}}$} \Comment{$\vec{\mathbb{D}}=D_1||D_2\ldots||D_k$}
\State $Loss[i]$=\Call{$\vec{\mathbb{M}}[i]$}{$Fed\_Batch$}
\EndFor
\EndFor
\State $Net\_Loss= \Sigma_{j=1}^{k}(W_i*Loss[j])$

\State $Net\_Loss.backward()$ \Comment{Back-propagate with Net\_Loss}
\EndFor
\EndProcedure
\Procedure{Specialized\_Training}{$\vec{\mathbb{M}}, \vec{\mathbb{D}}$}
\State $gradients(Trunk) \gets False$
\For{$i \in k$}
\For{$epoch \in Max\_epoch$}
\For{$ (data, label) \in \vec{\mathbb{D}}[i]$}
\State $Branch_i \gets loss(\vec{\mathbb{M}}[i](data), label).backward()$ \Comment{Backpropagation}
\EndFor
\EndFor
\EndFor
\EndProcedure
\Procedure{Model\_Creation}{$Trunk, Branch_1, Branch_2, \ldots, Branch_k$}
\State $Trunk \gets Arrangement(Conv, Relu, BatchNorm)$
\For{$ task \in total\_tasks$}
\State $Branch_{task} \gets Arrangement(Conv, Relu, BatchNorm, FC, Linear)$
\EndFor
\EndProcedure
\end{algorithmic}
\end{small}
\end{algorithm}

\textbf{(b) Dataset Preparation:}
TreeDNN requires joint training through multiple datasets. In order to maintain trade-off between bias and variance while performing training at each branch, it is required to design the minibatch considering all the datasets. Since minibatch is prepared with the participation from each of the datasets, we refer to it as $Fed\_Batch$ (federated batch). The preparation of the federated batch is shown in line:3, Algorithm 1.
\par
\textbf{(c) Training}
TreeDNN is trained in two steps: i) Generalized Training ii) Specialized Training.
\begin{itemize}
    \item \textbf{Generalized Training:} In first step, all the branches and trunk of the tree are trained together. During training, the samples are inferenced through the trunk and branches of the tree. This way, loss is calculated at each branch and then all the losses are aggregated as shown in equation \ref{training}. Cumulative loss is then backpropagated to the network through branch and the trunk equation \ref{backprop}. This step helps in training the trunk of the tree. Where $L$ is the loss value, $\hat{y}_j$ is the observed output at each branch and $y_j$ is the label at branch $j$.
    \begin{equation}
    \label{training}
        J(w^T,b)= \sum_{i=1}^{n}L_{Task:1}(\hat{y}_1^{i}, y_1^i) + \sum_{i=1}^{n}L_{Task:2}(\hat{y}_2^{i}, y_2^i) + \ldots+ \sum_{i=1}^{n}L_{Task:k}(\hat{y}_k^{i}, y_k^i)
        \end{equation}
    \begin{equation}
    \label{backprop}
        J(w^T,b).backward()
    \end{equation}
    
    \item \textbf{Specialized Training:} Now, it is required to tune each branch corresponding to a specific task. This step is performed by backpropagating the loss back to the network while keeping the gradient of trunk False equation \ref{f_grad}. Specialized training is applied at each branch independently as shown in equation \ref{specialized}. Where $I$ is the input samples.
    \begin{equation}
    \label{f_grad}
        \hat{v}=Trunk(I) \; , \; \; \; \; \;gradient(Trunk)= False\\ 
    \end{equation}
    \begin{equation}
    \label{specialized}
        \hat{y}_j= branch_j(\hat{v}) \; , \; \; \; \; \;
        Jj(w^T,b)= \sum_{i=1}^{n}L_{Task:j}(\hat{y}_j^{i}, y_j^i)
    \end{equation}

\end{itemize}

\begin{table}[t]
\caption{Case study of camera model, containing 8 DNN models}
\label{case_results}
\centering
\begin{tabular}{lll}
\hline
\textbf{Parameters}    & \textbf{Existing Solution} & \textbf{Proposed Solution} \\ \hline
\textbf{Memory}        & 120 MB                     & 68 MB                      \\ \hline
\textbf{Response Time} & 228 ms                     & 120 ms                     \\ \hline
\end{tabular}
\end{table}

\begin{table}[t]
\centering
\caption{Result of TreeDNN in comparison with dedicated backbone model}
\label{results}
\begin{tabular}{lllll}
\hline
\textbf{Tree}     & \textbf{Dataset} & \textbf{Samples} & \textbf{TreeDNN} & \textbf{SOTA (Mobilenet-V2)} \\ \hline
\textbf{Branch-1} & CIFAR10          & 50000         &  89\%  & 91\%          \\ \hline
\textbf{Branch-2} & CIFAR100         & 50000       &  66.9\%  & 68\%          \\ \hline
\textbf{Branch-3} & CALTECH101       &  9146     &  83.5\%   & 65\%          \\ \hline
\textbf{Branch-4} & CALTECH256       & 30,607       &  49.2\%  & 40\%          \\ \hline
\textbf{Branch-5} & SVHN             & 600,000       &   92.6\%  & 93.3\%        \\ \hline\end{tabular}
\end{table}


\subsection{Results and Discussion}
TreeDNN is developed with the agenda of solving the issues raised by loading and inferencing through multiple camera models or related systems. TreeDNN is a type of multi-task learning which supports the training through multiple datasets in a joint fashion. TreeDNN provides the flexibility of keeping the trunk of the tree on the working memory and because of that only a branch of the tree is required to be invoked at the time of execution. Hence, it reduces the loading time of the model, which can be witnessed in Table \ref{case_results}. The trunk of the tree works as the initial layers for any of the tasks and is shared among all the tasks therefore it helps in reducing the storage requirement of the initial layers' parameters.
Table \ref{case_results} showcases the case study of camera models and the performance of the TreeDNN when utilized in a real environment having 8 DNNs. The specified tasks for these models could be De-mosaic, Auto-Focus, Auto-Exposure, Auto-White balance, Auto-ISO, Color correction, Brightness enhancement etc. 
\par
The performance of the TreeDNN in the present experiment can be seen in Table \ref{results}. We have considered CIFAR10, CIFAR100, Caltech101, Caltech256, SVHN datasets to train the TreeDNN with 5 branches. SOTA results are the performance of MobileNet-V2 on aforementioned datasets when trained separately. It can be realized from the results that difficult datasets like Caltech256 and Caltech101 has shown improvement in performance because of utilizing the generalized trunk in the tree which works as regularization while training. However, there is a minute drop in the performance of relatively easy datasets.
TreeDNN can be deployed as a single model for all the tasks which also opens the direction of mounting new pretrained branch in an already grown TreeDNN to increase the scalability of the model for new tasks. Where the decision about selection of most appropriate branch can be taken by considering the task similarity.



\bibliographystyle{apalike}
\bibliography{biblio}
\includepdf[pages=-]{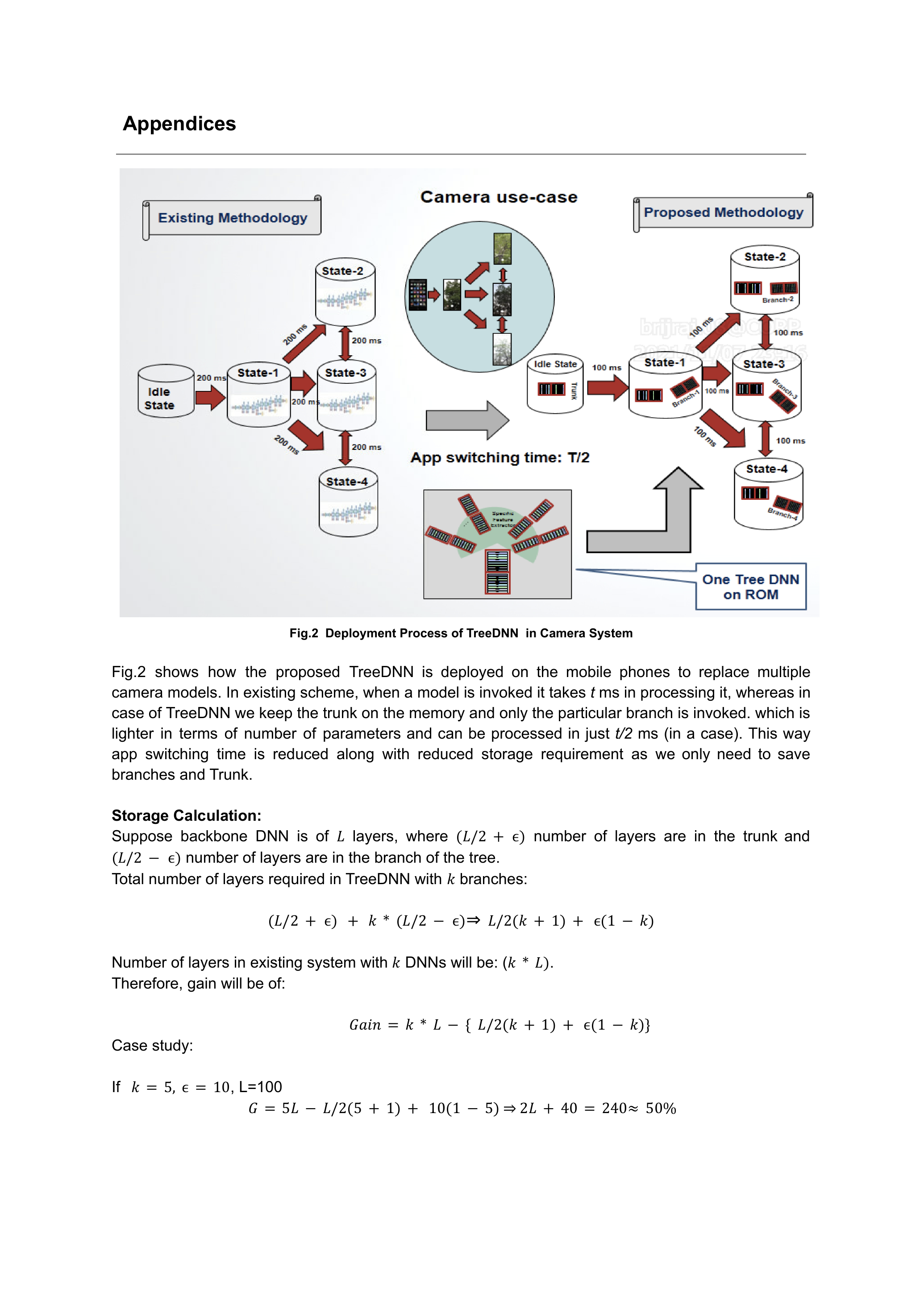}
\end{document}